\documentclass[conference]{IEEEtran}
\usepackage{tabularx,booktabs}
\usepackage{blindtext, graphicx}
\newcolumntype{C}{>{\centering\arraybackslash}X} 
\usepackage{lipsum}

\graphicspath{ {images/} }

\usepackage{cite}
\usepackage{float}
\usepackage{subcaption}
\usepackage[font={small,it}]{caption}
\usepackage{epstopdf}

\ifCLASSINFOpdf
\else
\fi

\begin{document}

\linespread{0.9}
\setlength{\parskip}{0cm}
\setlength{\parindent}{1em}
\setlength{\textfloatsep}{8pt}
%
\title{\huge Efficient Hardware Realization of Convolutional Neural Networks using Intra-Kernel Regular Pruning}

\author{\IEEEauthorblockN{Maurice Yang, Mahmoud Faraj, Assem Hussein, Vincent Gaudet}
\IEEEauthorblockA{Department of Electrical and Computer Engineering\\ University of Waterloo, ON, Canada\\
\{mm4yang, msfaraj, assem.hussein, vcgaudet\}@uwaterloo.ca}}


%


\maketitle

\begin{abstract}
The recent trend toward increasingly deep convolutional neural networks (CNNs) leads to a higher demand of computational power and memory storage. Consequently, the deployment of CNNs in hardware has become more challenging. In this paper, we propose an Intra-Kernel Regular (IKR) pruning scheme to reduce the size and computational complexity of the CNNs by removing redundant weights at a fine-grained level. Unlike other pruning methods such as Fine-Grained pruning, IKR pruning maintains regular kernel structures that are exploitable in a hardware accelerator. Experimental results demonstrate up to 10$\times$ parameter reduction and 7$\times$ computational reduction at a cost of less than 1\% degradation in accuracy versus the un-pruned case.
\end{abstract}


%
\IEEEpeerreviewmaketitle

\section{Introduction}\label{Introduction}
A Convolutional neural network (CNN) is a soft computing architecture that excels in prediction and pattern recognition. While neural networks have been studied for a wide range of applications \cite{cnn_speach}\cite{sugaya2016context}, a large focus of  CNN research targets 2-D image detection and recognition \cite{Krizhevsky}\cite{Lecun98gradient-basedlearning}. Recent CNN advances have popularized deeper network designs, with increasingly more layers and consequently larger models. Current state of the art designs comprise of millions of individual weights and require billions of computational operations for a single run. AlexNet, for example, requires over 200MB of memory for weight storage and 700Million FLOPs for inference\cite{Krizhevsky}. \vspace{2mm}

Currently GPUs are popularly chosen to run neural networks due to their high computational capacity\cite{kisavcanin2017deep}. Although powerful, GPUs suffer from high power consumption and a bulky footprint, making them unsuitable for energy-critical mobile CNN designs. Alternatively, custom hardware designs are compact and can achieve high power efficiency, offering a promising solution for these portable applications. Because of this, there is research interest in implementing CNNs in VLSI\cite{Chen:2014:DMS:2742155.2742217} or using FPGAs\cite{optimalFPGA}\cite{eyeriss}\cite{openclfpga}. Existing hardware designs have demonstrated that while arithmetic operations can be executed with low energy consumption, memory access can pose a significant bottleneck in terms of energy efficiency. This is because modern CNN models are often too large to be fit into on-chip memory and must be stored instead on DRAM. It is shown in \cite{deepcompression} that for a 45nm CMOS process, a single 32-bit DRAM access can consume 100 times more energy than a single 32-bit SRAM access and 2000 times more energy than a single 32-bit floating point multiply.  \vspace{2mm}

Reducing the memory and computational requirements for running neural networks is an active area of research. Weight quantization\cite{limitedprecision} is commonly employed to reduce the resolution of weight parameters down to fixed-point or integer levels. Corresponding hardware designs benefit from lower memory requirement and simpler computational hardware. Stochastic Computation\cite{stochastic} and Network Binarization\cite{binaryconnect} are other promising techniques that significantly lower the hardware and energy cost for arithmetic operations. \vspace{2mm}

Other researchers have explored removing model parameters to reduce network size. Sparsity regularization\cite{groupbraindamage} is used during training to incentivize certain weights towards being zero valued; since zero-valued weights contribute nothing to the output, they can be effectively ignored. Similarly, connection pruning\cite{connectionpruning} can be applied on a conventionally trained network to remove unimportant weights. For both techniques, the resulting network is denoted as sparse since kept weights are scattered throughout the network model. Previous researchers employed fine-grained pruning\cite{learnweightsconnections} to remove individual weights, achieving a theoretical 9 times memory and 3 times computational reduction on AlexNet without loss in accuracy. While fine-grained pruning is proven to reduce network size, the irregular weight distribution of the resulting sparse model makes it difficult to attain practical savings. \cite{deepcompression} tackles this by representing sparse network models in Compressed Sparse Row (CSR) format, where only non-zero weight values and their respective locations are stored, allowing for dense storage of highly irregular structures. Utilizing fine-grained pruning along with weight quantization, weight sharing and Huffman Encoding, \cite{EIE} was able to store large CNN models solely on SRAM. Coarse-grained pruning\cite{exploringregularsparsity} was proposed as an alternative pruning method, where entire vectors, kernels\cite{compactDCNNfilter} or filters\cite{pruningweights} are removed. Although pruning at coarser granularities generates more structured models with better data locality, it is more destructive and does not achieve the same performance as fine-grained pruning\cite{exploringregularsparsity}. \cite{anwar2017structured} presented the concept of intra-kernel strided structured sparsity, which prunes in accordance to rigid structural constraints. While this idea leads to promising hardware design, the imposed restrictions are harsh and lower achievable sparsity. \cite{gross} explored activation pruning using random masks generated from Linear Shift Feedback Registers. While this method is appealing due to the low hardware cost overhead, the network accuracy degradation from pruning is potentially high since all activations are equally susceptible to removal regardless of their importance.  \vspace{2mm}

The objective of this research is to reduce memory and computational cost for CNN inference by proposing an Intra-Kernel Regular (IKR) pruning scheme that uses generated pruning patterns to preserve important weights while eliminating insignificant weights at the intra-kernel level. Our approach reaps the benefits from fine-grained pruning while maintaining predictable kernel patterns that can be exploited using specialized hardware. Moreover, the resulting sparse kernels can be stored very compactly in compressed sparse pattern (CSP) format, a representation that exclusively keeps non-zero weights and the corresponding mask index. The generation and selection of pruning patterns are also contributions of this paper. \vspace{2mm}

This paper is divided into five Sections. In Section~\ref{Intra-Kernel Regular Pruning}, the IKR pruning scheme is described in detail along with background information on the CNN operation. Section~\ref{Sparse Computation in Hardware} reviews the hardware architecture for inference on IKR sparse networks. The simulation environment is described and results are reported in Section \ref{Simulation and Results}. Finally, Section~\ref{Conclusion} provides concluding remarks and discussion of future research direction.

\section{Intra-Kernel Regular Pruning}\label{Intra-Kernel Regular Pruning}
IKR pruning structurally eliminates weights at an intra-kernel level while retaining original accuracy. The proposed scheme supports pruning in the convolutional and Fully Connected (FC) layers; however, for the sake of simplicity we clarify the methodology in terms of the convolutional layer only. Prior to pruning, a neural network is conventionally trained and is set as the baseline. The trained network model is extracted and kernels with similar locality are grouped into sets. We define a network with $m$ layers, such that the set of layers is $L=\{l_{1},l_{2},\dots,l_{m}\}$. The $\ell$-th layer, $l_{\ell}$, has $N_{sets}^{\ell}$ sets of kernels such that  $l_{\ell}=\{S_{1}^{\ell},S_{2}^{\ell},\dots, S_{N_{sets}^{\ell}}^{\ell}\}$. Each set of kernels $S_{i}^{\ell}$, where $i=1,2,\dots,N$, includes $N_{ker}^{\ell}$ kernels such that $S_{i}^{\ell}=\{W_{1},W_{2},\dots,W_{N_{ker}^{\ell}}\}$. The $j$-th kernel belonging to $S_{i}^{\ell}$ is denoted as $W_{i,j}^{\ell}$.\vspace{2mm}

Pruning patterns indicate the locations at which parameters should be kept or eliminated. When pruning at fine-granularity, these patterns are applied at the kernel level, specifying the individual weights that should be removed. The resulting kernel structure is described as irregular since the locations of kept weights are random. Similarly, IKR pruning operates at a fine-grained level; however, we challenge irregularity by memorizing the specific pruning pattern applied to each kernel, allowing us to recall the exact location of kept weights. To reduce storage costs, we impose a restriction on the number of possible pruning patterns. Specifically, for each $S_{i}^{\ell}$, we have $N_{pat}^{\ell}$ possible pruning patterns, $C_{i}^{\ell}=\{ p_{1},p_{2}, \dots, p_{N_{pat}^{\ell}} \}$. A pattern belonging to $C_{i}^{\ell}$ is denoted as $p_{i,k}$, where $k=1,2,\dots,N_{pat}^{\ell}$.\vspace{2mm}

\begin{figure}[t]
\centering
\includegraphics[scale=0.52]{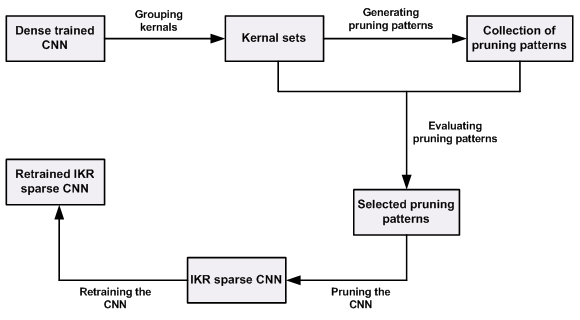}
\caption{Schematic depiction of IKR scheme}
\label{fig:sdikrs}
\end{figure}

The objective of pruning is to maximally reduce the number of parameters in the network model while suffering minimal network damage; therefore, it is vital for pruning patterns to retain important weights. We gauge the suitability of a pattern $p_{i,k}$ to a kernel $W_{i,j}$ using the quality metric, $Q(p_{i,k}, W_{i,j})$, and use the highest quality pattern-kernel pair during pruning. This process is explained in more details in Section~\ref{Mask Pattern Generation}. The resulting sparse model is retrained to regain the baseline accuracy.  Fig. \ref{fig:sdikrs} illustrates the mechanism for the IKR pruning. \vspace{2mm}

Pruning in the FC layer follows the same methodology that is formerly outlined. The preface for IKR pruning involves grouping kernels into sets. Although connections in the FC layer are instead represented by a matrix of individual weights, kernels can be artificially created. For example, by grouping 16 parameters, a $4\times4$ kernel is formed. The IKR pruning follows naturally thereafter. 

\begin{figure}[t]
\centering
\includegraphics[scale=0.52]{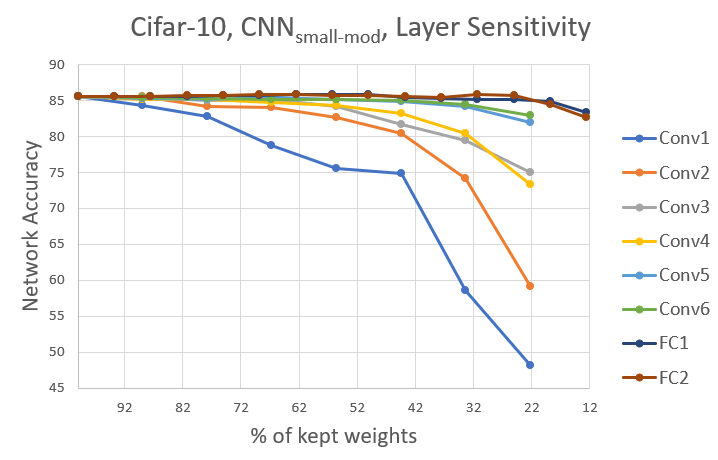}
\caption{Sensitivity to pruning of Convolutional and FC layers from $CNN_{small}$}
\label{fig:layersensitivity}
\end{figure}

\subsection{CNN Computation}\label{CNN Computation} 
In a typical CNN, the most computationally intensive operations reside in the convolutional and the FC layer. The convolutional layer receives $n_{in}$ input feature maps and produces $ n_{out}$ output feature maps. Connections between input and output are represented by $ n_{out}$ filters, each of which has $ n_{in}$ kernels of dimensions $K \times K$. The convolutional layer performs convolutions between input feature maps and kernels to generate output feature maps, as shown in (\ref{eq:conv}), where $ f_{i}^{out}$ denotes the $i$-th output feature map and $ f_{j}^{in} $ denotes the $j$-th input feature map. It is observed that convolutional layers occupy a majority of the required computations in a CNN. \vspace{2mm}
\begin{equation}
\label{eq:conv}
f_{i}^{out} =\sum_{j}^{n_{in}}  f_{j}^{in} * W_{i,j} + b_{i}
\end{equation}

The FC layer has all to all connections between the input and the output feature maps, which can be represented as a vector-matrix multiplication between the input and the weights. This operation is summed up in (\ref{eq:fc}). Although less computationally demanding than the convolutional layer, the FC layer contains the most weights and thus requires high memory bandwidth for operation. 

\begin{equation}
\label{eq:fc}
f^{out} = W \cdot f^{in} + b
\end{equation}

\begin{figure}[t]
\centering
\includegraphics[scale=0.52]{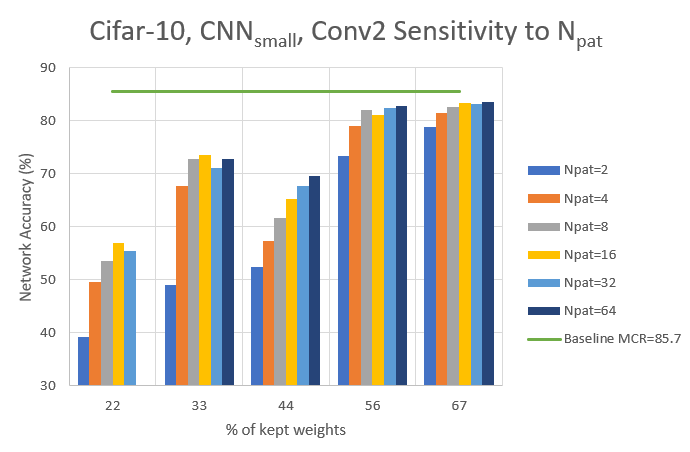}
\caption{The affect of $N_{pat}^{2}$ on the accuracy of $CNN_{small}$ at various sparsity}
\label{fig:conv2sensitivity}
\end{figure}

\subsection{Mask Pattern Generation}\label{Mask Pattern Generation}
Pruning severs connections within the CNN, reducing the number of learnable parameters and damaging its ability to correctly perform classification. A crucial step during pruning involves determining which parameters can be removed with least affect on network performance. Previous research \cite{learnweightsconnections}\cite{pruningweights} assigns the importance of a weight to its magnitude. Consequently, pruning patterns that retain a high absolute summation are characterized as having high quality. Alternatively, \cite{groupbraindamage}\cite{compactDCNNfilter} assess pruning patterns by first applying the pattern and then evaluating the drop in misclassification rate (MCR) on the validation set. Patterns resulting in the smallest MCR drop are considered to be least damaging. Since both methodologies produce comparable performance, the magnitude-based approach is adopted in this paper as it is simpler. \vspace{2mm}

\begin{figure}[t]
\centering
\includegraphics[scale=0.48]{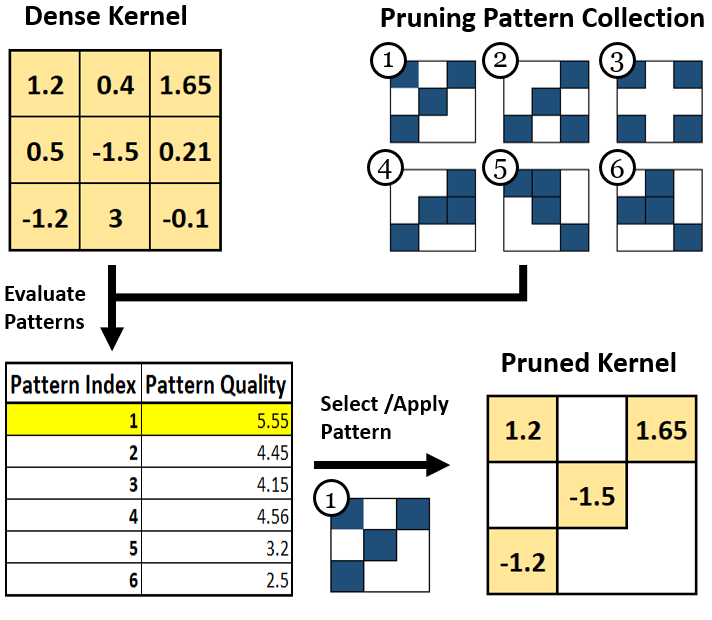}
\caption{Selecting a pruning pattern based on quality}
\label{fig:selectmask}
\end{figure} 

Each pruning pattern is represented by a mask of the same shape as the kernels it is targeting. Elements within the mask are either zero-valued or one-valued, with zero representing a prune and one representing a keep. A mask is applied to a kernel via elementwise matrix multiplication, producing a masked kernel. The suitability of a pruning pattern $p_{i,k}$ to a kernel $ w_{i,j}$ is determined by the quality metric $Q(p_{i,k}, w_{i,j})$, which is expressed in (\ref{eq:quality}). The highest quality pattern for the kernel $W_{i,j}$ is found by an exhaustive search through $C_{i}^{\ell}$, as illustrated in Fig. \ref{fig:selectmask}. During pruning, the pattern is permanently applied by overwritting the original kernel with the masked kernel. In consideration of the hardware, equal pruning is enforced, where each pruning pattern in layer $\l_{\ell}$ keeps the same number of weights $N_{keep}^{\ell}$. 

\begin{equation}
    \label{eq:quality}
Q(p_{i,k}, W_{i,j}) =\sum  \left| p_{i,k} \odot W_{i,j} \right|
\end{equation}

\begin{figure*}
\begin{subfigure}{0.3\textwidth}
\includegraphics[width=\linewidth]{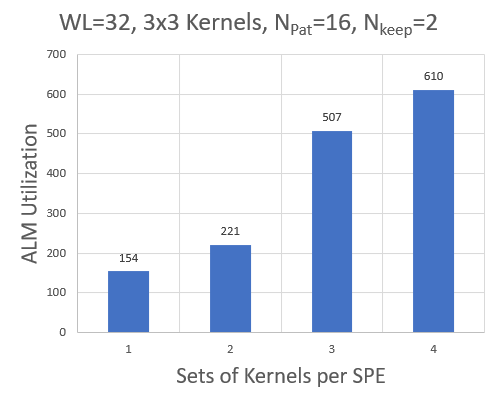}
\caption{Varying set coverage.} \label{fig:paramcover}
\end{subfigure}
\hfill%
\begin{subfigure}{0.33\textwidth}
\includegraphics[width=\linewidth]{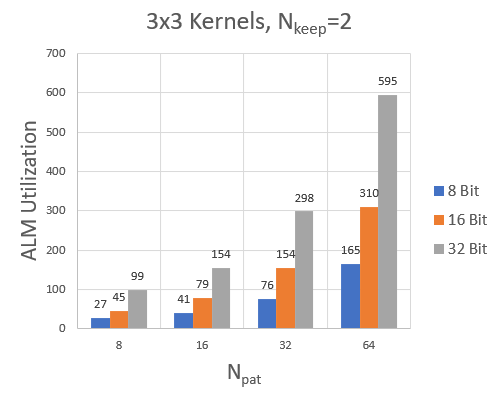}
\caption{Varying $N_{pat}$ for 8-bit, 16-bit and 32-bit word lengths.}  \label{fig:paramker}
\end{subfigure}
\hfill%
\begin{subfigure}{0.3\textwidth}
\includegraphics[width=\linewidth]{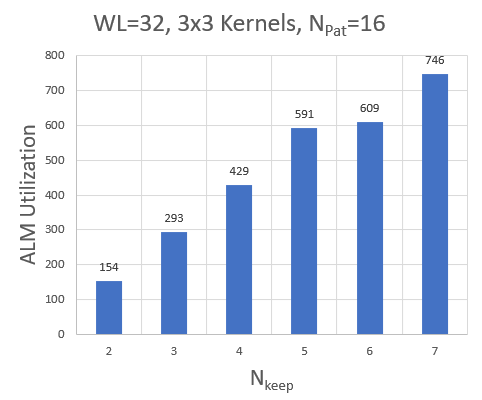}
\caption{Varying $N_{keep}$.} \label{fig:nkeep}
\end{subfigure}
\caption{ALM utilization for the pattern selector module, with respect to an increasing demand for (a) set coverage, (b) pattern coverage and (c) number of kept weights.} \label{fig:paramkeep}

\end{figure*}

Pruning pattern collections are populated through a candidate selection process. Ten “promising” pruning patterns are generated for each kernel in $S_{i}^{\ell}$, each of which retains a different permutation of top valuable weights. These patterns are potential candidates for inclusion into $C_{i}^{\ell}$. It should be noted that although a pattern may be suitable for a particular kernel $ W_{i,j}$, it may not suit other kernels in $S_{i}^{\ell}$. Since the population of $C_{i}^{\ell}$ is limited to only $N_{pat}^{\ell}$, candidates with the best representation of $S_{i}^{\ell}$ should be chosen. From the entire set of promising pruning patterns generated from $S_{i}^{\ell}$, $N_{pat}^{\ell}$ candidates with the highest overall quality are selected to populate $C_{i}^{\ell}$.

\begin{table*}
\center
 \caption{Architecture of LeNet-5 and $CNN_{small}$}
\label{architecture}
\begin{tabular}{|c|c|c|c|}
\hline
Network     & Architecture & DataSet & Baseline MCR \% \\ 
\hline
LeNet-5   & 1x20C5-MP2-1x50C5-MP2-500FC-10Softmax      & MNIST          & 0.6    \\ 
\hline
$CNN_{small}$   & 2x128C3-MP2-2x128C3-MP2-2x256C3-256FC-10Softmax      & CIFAR-10          & 14.3    \\ 
\hline
\end{tabular}
\end{table*}

\subsection{Layer Sensitivity}\label{Layer Sensitivity}
Pruning on each layer of the CNN has a different impact on the network performance. Certain layers are tolerant to weight removal and can achieve high sparsity without significant loss in accuracy, while others are more sensitive. Following the approach in \cite{pruningweights}, we investigate the sensitivity of each layer to pruning. Starting with an original dense model, each layer is isolated and pruned with incrementally higher degree of sparsity, and validation accuracy is recorded at every step. Based on observed sensitivity, we empirically choose how aggressively each layer is pruned by choosing the number $N_{keep}^{\ell}$. For example, sensitive layers are chosen to have a larger $N_{keep}^{\ell}$. Fig.~\ref{fig:layersensitivity} shows the network accuracy as each layer is individually pruned for $CNN_{small}$ (i.e., the $CNN_{small}$ is a VGG16 inspired CNN containing 6 convoutional and 2 FC layers operating on the CIFAR-10 dataset, adopted from \cite{compactDCNNfilter}). It is observed that accuracy suffers the most when pruning the first two stages. To explore the impact of $N_{pat}$ on network accuracy, simulation is conducted using the second convolutional layer of $CNN_{small}$ as a reference. With the other layers untouched, MCR is measured for various values of $N_{pat}^{2}$ at various sparsity.  It can be observed from Fig. \ref{fig:conv2sensitivity} that increasing $N_{pat}^{2}$ beyond the value of 8 gives diminishing returns. 

\subsection{Storing Sparse Matrices}\label{Storing Sparse Matrices}
To obtain practical savings, the resulting sparse matrices must be stored in a dense format. \cite{deepcompression} stores the sparse matrices using Compressed Sparse Row (CSR) notation, a representation that only keeps non-zero weights and their respective indices. We propose a similar format called Compressed Sparse Pattern (CSP) to store IKR sprase kernels. Leveraging the fact that 1) kernels within the same layer keep the same number of weights after pruning, 2) pruning patterns determine the locations of kept weights within each kernel and 3) only $N_{pat}^{\ell}$ pruning patterns are accessible for each kernel within  $S_{i}^{\ell}$, CSP exclusively keeps non-zero weights and the corresponding mask pattern index. The number of bits required to represent the pattern index is equal to $\log_2 N_{pat}^{\ell}$.

\section{Sparse Computation in Hardware}\label{Sparse Computation in Hardware}
It is difficult to exploit irregular intra-kernel sparsity in hardware since the locations and the number of non-zero weights vary between kernels. As previously mentioned, \cite{deepcompression} challenged irregular sparsity by storing non-zero weights in CSR format, transforming irregular structures into regular representations. We propose an alternative approach, where we prune with regularity in mind. IKR pruning restricts the variability in the composition of kernels because the number of pruning patterns is limited. Futhermore, by storing kernels in CSP format, exact composition of every kernel is known. It is expected that IKR sprase networks can be implemented efficiently using specialized hardware resembling designs that exist in the literature. 

\begin{figure}[t]
\centering
\includegraphics[scale=0.4]{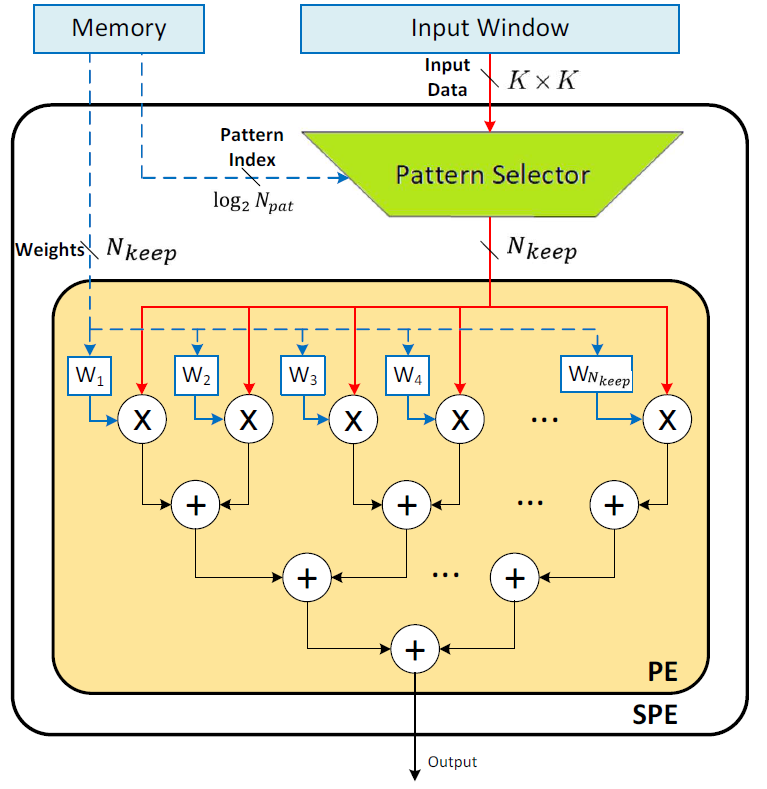}
\caption{Block diagram of the SPE architecture}
\label{fig:SPE}
\end{figure}

\subsection{Sparse Processing Engine}\label{Storing Processing Engine}
In CNN accelerators, the Processing Engine (PE) is a core building block. Past research commonly adopted a PE design consisting of multipliers and an adder tree \cite{optimalFPGA}. The purpose of the PE is to perform inner product operations between a sliding input window and a kernel matrix. It is common practice to tile PEs for parallel computation and to increase throughput; however, the extent of tiling may be restricted due to hardware resource limitations. Qualitatively, we propose a Sparse Processing Engine (SPE) design for IKR sparse networks to achieve the same functionality as conventional PEs but at a potentially lower resource cost. \vspace{2mm}

Fig. \ref{fig:SPE} shows an overview of the SPE architecture. SPE is a modification on the conventional PE structure, containing a small conventional PE and a pattern selector block. Unlike the conventional PE, SPE operates on IKR sparse matrices, where each individual SPE block is designed to operate uniquely on one set of kernels. Computation on pruned elements are redundant, so the SPE computes the inner product only between non-zero kernel weights and the corresponding input data. Since each SPE operating on $S_{i}^{\ell}$ requires only $N_{keep}^{\ell}$ multipliers and an adder tree of depth $\log_2 N_{keep}^{\ell}$, the reduction in arithmetic units is proportional to network sparsity. The pattern selector is a collection of multiplexors that chooses input data based on the selected pruning pattern. Hence, the inherent trade-off of using SPE is less cost in terms of arithmetic units for an extra overhead in logic. We argue that this compromise can be justified in FPGA designs as on-chip DSP units are scarce while logic fabric is plentiful. \vspace{2mm}

While the SPE is proposed to operate solely on one set of kernels, it can be modified to cover multiple sets by adding extra logic to the pattern selector. This alternate design is advantageous if the number of unutilized LUTs is high while DSP count is low. By generalizing each SPE over a larger set of kernels, less SPEs are required for each layer. In essence, the trade-off is less DSP utilization for an increased cost of logic. Fig. \ref{fig:paramcover} summarizes the Adaptive Logic Module (ALM) utilization of pattern selector designs with respect to the increasing coverage. The design is compiled for Cyclone V GX (5CGXFC9E7F35C8) using the Altera Quartus Prime software. It is observed that the ALM utilization scales proportionally with set coverage. \vspace{2mm}

Having a large population of SPEs corresponds to high parallelism; however, if too many SPEs are specified, the hardware implementation may not be feasible. For layer $l_{\ell}$, $N_{sets}^{\ell}$ SPEs are required. Since the cost of each SPE is influenced by $N_{keep}^{\ell}$, $N_{pat}^{\ell}$ and the targeted kernel size, a design space exploration must be conducted to find the optimal settings in relation to a particular platform and network. Fig. \ref{fig:paramker} depicts how ALM utilization is affected by $N_{pat}$ and word length; it is observed that the ALM cost is low if $N_{pat}$ and word length are kept low. Fig. \ref{fig:nkeep} shows how ALM utilization changes corresponding to $N_{keep}$.

\subsection{Other Optimizations}\label{Other Optimizations}

The freedom granted by IKR pruning enables further hardware optimizations. By choosing $N_{keep}$ to be a power of two, we can achieve a balanced adder tree structure, an idea previously explored in  \cite{hardwarefriendly}. If the adder tree is unbalanced, extra flip flops are required to buffer the peripheral inputs and the tree depth must be increased. A balanced adder tree requires no extra flip flops. In addition, $N_{keep}$ can be chosen to achieve higher memory bandwidth utilization.  The detailed explaination of this concept can be found in \cite{hardwarefriendly}.

\begin{table}[t]
 \caption{Parameters used during IKR pruning on LeNet-5.}
\label{params_lenet}
\begin{tabular}{|c|c c c c|}
\hline
Layer    							&Kernel Size   &$N_{sets}$ &$N_{pat}$	 &$N_{keep}$		\\ 
\hline
C1$(1\times20)$	   				& 5$\times$5 	 &2			&8    	&6  	\\
\hline
C2$(20\times50)$					& 5$\times$5 	 &10       	       &8		&3\\
\hline
FC$(800\times500)$					& 5$\times$5 	 &10             	&16		&2\\
\hline
FC$(500\times10)$					& 5$\times$5 	 &5            	&16		&2\\
\hline
\end{tabular}
\end{table}

\begin{table}[t]
 \caption{Parameters used during IKR pruning on $CNN_{small}$.}
\label{params_small}
\begin{tabular}{|c|c c c c|}
\hline
Layer    							&Kernel Size   &$N_{sets}$ &$N_{pat}$	 &$N_{keep}$		\\ 
\hline
C1$(3\times128)$	   				& 3$\times$3 	 &3			&16    	&6  	\\
\hline
C2$(128\times128)$					& 3$\times$3 	 & 8       	       &16		&3\\
\hline
C3$(128\times128)$					& 3$\times$3 	 & 8              	&16		&2\\
\hline
C4$(128\times128)$					& 3$\times$3 	 & 8              	&16	 	&2\\
\hline
C5$(128\times256)$					& 3$\times$3 	 & 16              	&16	 	&2\\
\hline
C6$(256\times256)$					& 3$\times$3 	 & 16            	&16	 	&2\\
\hline
FC$(256\times256)$					& 4$\times$4 	 & 8              	&16		&3\\
\hline
FC$(256\times10)$					& 4$\times$4 	 & 5             	&16		&4\\
\hline
\end{tabular}
\end{table}

\begin{table*}[h]
\centering
\fontsize{8}{8}\selectfont
 \caption{Pruning statistics for LeNet-5 and $CNN_{small}$. FG: Fine-Grained, FMK: Feature Map followed by Kernel}
\renewcommand\arraystretch{1.3}
\label{PruneRatio}
\begin{tabular}{|m{2.7cm}|>{}m{0.9cm}|m{0.8cm}|m{1cm}|m{1cm}|m{1.5cm}|m{1.5cm}|}
\hline
Pruned Network		  					& Baseline Error  	& \centering Final  Error		&Weights		& Weight Density 		&Computations		& Computational Density 		 \\ 
\hline
\bfseries LeNet-5 IKR						&0.6\%			&1.1\%					&42.7K			&10\% 	 			&63.4K				&13.8\%     		\\
LeNet-5 FG \cite{learnweightsconnections} 		&0.8\%			&0.77\%					&34.5K			&8\%				&73.3K				&16\%		\\
\cline{1-5}

\hline

\bfseries \boldmath$CNN_{small}$	IKR				&14.3\%			&15.2\%		&390K	&23.1\% 	 	&145M		&15.3\%     		\\
$CNN_{small}$  FMK \cite{compactDCNNfilter}			&16.26\%			&17.26\%		&- 		&25\%		&- 			&-	\\
\cline{1-7}
\end{tabular}
\end{table*}

\section{Simulation and Results}\label{Simulation and Results}
 
To investigate the performance of the IKR pruning, simulations were conducted in python using TensorFlow. The IKR pruning was applied to two different CNNs, namely LeNet-5, which is introduced in \cite{Lecun98gradient-basedlearning}, and $CNN_{small}$. The architectures of the two networks are outlined in Table \ref{architecture}. In parallel with  \cite{compactDCNNfilter}, we follow a similar notation for describing network architecture. In Table \ref{architecture}, 2x128C3 denotes two adjacent convolutional layers having 128 feature maps each and the kernels are of dimensions 3 x 3. MP2 denotes one non-overlapping max pooling layer with dimensions 2 x 2 and stride 2. 256FC denotes an FC layer with 256 output nodes. 10Softmax denotes 10 nodes with SoftMax regression. Dropout \cite{dropout} is applied after each MP2 layer with 50\% keep probability to prevent overfitting. The networks were trained using Stochastic Gradient Descent (SGD) and Adam optimization with mini-batches of 128 images using 32-bit floating point numbers. For each layer $l_\ell$, the parameters $N_{keep}^{\ell}$, $N_{set}^{\ell}$and $N_{pat}^{\ell}$ are empirically chosen to balance network accuracy and a feasible hardware implementation. The parameters used during IKR pruning of LeNet-5 and $CNN_{small}$ are reported in Table \ref{params_lenet} and Table \ref{params_small} respectively.

\subsection{LeNet-5 on MNIST}
The similation tests were carried out on LeNet-5, comparing the IKR pruning scheme to Fine-Grained Pruning \cite{learnweightsconnections} in terms of weight and computational density. Weight density refers to the number of weights in the pruned network as a percentage of the baseline network; computational density signifies the number of multiplication/addition operations required for one forward pass of the pruned network as a percentage of the baseline. The evaluation of the two techniques was performed on the MNIST dataset. MNIST is a collection of 28 x 28 greyscale images, with each image containing a single handwritten digit from 0 to 9. We divided the original training set of 60,000 samples into a 55,000 sample training set and a 5,000 sample validation set.  Then, the random contrast and random flip transformations are applied to the replicated images. The CNN was trained for 15 epochs and then for 10 epochs using learning rates of 0.001 and 0.0001 respectively. During retraining, the learning rate was set at 0.005 for 10 epochs and 0.0001 for 10 epochs. Table \ref{PruneRatio} shows that the IKR pruned network retains 10\% of the weights and 13.8\% of the computations of the baseline network, corresponding to a 10 times network compression and a 7 times computational reduction. As seen in the table, the IKR pruning achieves comparable results compared to Fine-Grained pruning. The final network error rate is 0.5\% higher than the baseline error-rate.

\subsection{$CNN_{small}$ on CIFAR-10 }
$CNN_{small}$ is used to perform classifcation on the CIFAR-10 dataset. The simulation tests compare the IKR pruning to the Feature Map followed by Kernel-Level (FMK) pruning \cite{compactDCNNfilter} in terms of weight density and accuracy. CIFAR-10 is a collection of 32 x 32 RGB images, with each image belonging to one of 10 object classes, such as cat, frog, airplane, etc.  We divided the original training set of 50,000 samples into a 45,000 sample training set and a 5,000 sample validation set. Prior to training, the input images are preprocessed with a whitening transformation. To artificially double the number of training images, the training set is duplicated; and random contrast and random flip transformations are applied to the replicated images. The CNN was trained for 50 epochs and then for 20 epochs using learning rates of 0.001 and 0.0001 respectively. During retraining, the learning rate was set at 0.001 for 20 epochs and 0.0001 for 20 epochs. Table \ref{PruneRatio} shows that the IKR pruning compresses the original network size by 4 times and reduces the required computations by 6 times. It can be seen from the table that with the same 1\% accuracy degradation, the IKR pruning achieves slightly higher weight reduction compared to the FMK pruning. While the FMK pruning did not provide computational savings, it is reported for the IKR pruning.

\section{Conclusion}\label{Conclusion}
This paper has tackled structured pruning of CNNs for efficient hardware implementation. An IKR pruning scheme was proposed to compress CNNs at fine granularity while maintaining regular kernel structures. The design of a sparse processing engine, namely SPE, was proposed to operate on the IKR pruned CNNs. By applying the IKR pruning to two benchmark CNNs, LeNet-5 and $CNN_{small}$, using two different datasets, it has been demonstrated that the IKR pruning scheme achieves comparable accuracy and sparsity as compared to Fine-Grained and Kernel-Level pruning. The future direction of this research will focus on efficient implementation of the proposed CNN in hardware.



%

\bibliographystyle{IEEEtran}
\small{
	\bibliography{references}
}

\end{document}